\documentclass{article}

\usepackage[cloudcomputing, nonatbib, final]{cosparws_abs} 

\usepackage[utf8]{inputenc} 
\usepackage[T1]{fontenc}    
\usepackage{hyperref}       
\usepackage{url}            
\usepackage{booktabs}       
\usepackage{amsfonts}       
\usepackage{nicefrac}       
\usepackage{microtype}      

\usepackage{graphicx}
\usepackage{xcolor}

\title{SpaceML: Distributed Open-source Research with Citizen Scientists for the Advancement of Space Technology for NASA}

\author{%

     \textbf{Anirudh Koul} \thanks{Equal Contribution} \\
  \and
  
 \textbf{ Siddha Ganju}  \footnotemark[1]  \\
  \and

     \textbf{Meher Kasam} \footnotemark[1] \\
    \and
  
  \textbf{James Parr}  \\
}
    
\begin{document}

\maketitle

\begin{abstract}

Traditionally, academic labs conduct open-ended research with the primary focus on discoveries with long-term value, rather than direct products that can be deployed in the real world. On the other hand, research in the industry is driven by its expected commercial return on investment, and hence focuses on a real world product with short-term timelines. In both cases, opportunity is selective, often available to researchers with advanced educational backgrounds. Research often happens behind closed doors and may be kept confidential until either its publication or product release, exacerbating the problem of AI reproducibility and slowing down future research by others in the field. As many research organizations tend to exclusively focus on specific areas, opportunities for interdisciplinary research reduce. Undertaking long-term bold research in unexplored fields with non-commercial yet great public value is hard due to factors including the high upfront risk, budgetary constraints, and a lack of availability of data and experts in niche fields. Only a few companies or well-funded research labs can afford to do such long-term research. With research organizations focused on an exploding array of fields and resources spread thin, opportunities for the maturation of interdisciplinary research reduce. Apart from these exigencies, there is also a need to engage citizen scientists through open-source contributors to play an active part in the research dialogue. We present a short case study of SpaceML, an extension of the Frontier Development Lab, an AI accelerator for NASA. SpaceML distributes open-source research and invites volunteer citizen scientists to partake in development and deployment of high social value products at the intersection of space and AI.
\end{abstract}

\section*{Introduction}

Increasingly accessible satellite data, a booming global space industry, and an abundance of statistical and newer AI technologies opens up the potential to accelerate interdisciplinary scientific research. This comes with the challenges of scaling traditional scientific processing workflows and, more importantly, maintaining the quality of deployed AI, which is plagued by issues of reproducibility, management of knowledge transfer, and lack of expertise. NASA Science Mission Directorate has declared ~\cite{smd} a commitment to open science, with an emphasis on continual monitoring and updating of deployed systems, improved engagement of the scientific community with citizen scientists, and data access to the wider research community for robust validation of published research results. SpaceML is one such international AI accelerator that applies startup and agile software development principles to research, thereby reducing time to success significantly.

The SpaceML~\footnote{http://spaceml.org/} open-source research program takes eligible projects from the Frontier Development Lab (FDL)~\footnote{FDL is an AI accelerator supported by both NASA and the European Space Agency, along with a few other corporate and government partners.}\cite{Ganju2020FDL} and enables continued development of those projects beyond pure research, to develop and deploy solutions usable in the real world, eventually handing them over to stakeholders. In addition to the core goals of the project,  extensions to the project for “nice to have” capabilities are available as mini-projects to open-source contributors. Mentorship support from industry experts and cloud computing resources are made available to the contributors where necessary. SpaceML's first iteration in 2020 converted many high school students to citizen scientists with unique stories, like a citizen scientist from Nigeria, who improved FDL’s advanced AI-powered meteor detector pipeline~\cite{Zoghbi2017SearchingFL, JENNISKENS201821} for the NASA CAMS~\cite{JENNISKENS201140} project along with an interactive interface in his local cyber-cafe.

FDL and SpaceML undertake interdisciplinary research with researchers and experts from different organizations, yet operate as one team. This ensures a smooth handover and efficient transfer of knowledge and capabilities, reducing transitional failures, ultimately resulting in robust and reproducible work.

\medskip

\section*{Key Features of SpaceML}
\label{sec:SpaceMLtimeline}
We focus on how SpaceML utilizes cloud computing technologies, open-source challenges, and volunteer citizen scientists to develop industry-level AI technologies measured by TRL4ML~\cite{lavin2020technology}.

\textbf{Open-source Challenges}: As a first step, eligible FDL challenges are brought under the umbrella of SpaceML. FDL researchers create a set of miniature extension projects on a scale of easy to hard. Most of these mini-projects have a very narrow focus and are meant to be built by a competent developer with three weeks of effort. The mini-projects are well defined to include availability and access of data, a predefined technology stack, reference material (research papers, web articles, tutorial videos, etc.), and potential solution ideas. Global calls for SpaceML volunteer open-source contributions invite people from different backgrounds and different countries for the positions of researchers and domain leads. Access to experts in various fields is made possible through FDL and SpaceML public-private partnerships. Once the contributions reach a certain level of maturity, new volunteer mentors through the FDL and SpaceML network join in to help boost the project further.
{Regular weekly and gated expert reviews} provide direct and actionable feedback, and serve as key decision points to {measure the correctness, impact, and relevance of research}.

\textbf{Cloud Compute}: To reduce the barrier to entry, FDL researchers package accessible code and instructions on GitHub, Google Colab (with free GPUs) so that the new contributors can run it with relative ease. Colab notebooks allow you to combine executable code and rich text in a single document, along with images, HTML, LaTeX thus enabling code execution and instructions or information to co-occur. Colab has absolute zero configuration or setup required allowing the open-source contributors to get up and running quickly. Colab also offers free access to advanced computing hardware like GPUs or TPUs, which significantly accelerate development. The GitHub repository also contains a video explaining the overall vision of each project and descriptions of project-specific beginner challenges, along with detailed guidance on how to get started. The beginner challenges, akin to an interview, allow competent and motivated contributors to move further in the challenge. At the same time, for people who are not yet familiar with the technical stack (science or AI), this serves as an educational opportunity to level up with the help of the detailed guides. Beginner guides can include ``How to get started in AI'', ``How to use GitHub effectively'', ``How to access satellite imagery data'', ``Code Quality standards'', ``Citation guide'', and, ``Code Reproducibility standards''. Contributors who have shown enough progress are then provided with Google Cloud credits to scale up their AI training and storage needs. 

\textbf{Mentorship}: FDL researchers, who worked on the original project, are now the stakeholders of the open-source project, and switch their roles to mentor while providing weekly mentorship. Throughout the process of improving the project, the contributors eventually become an expert in the niche area of research.  Once the contributions reach some maturity, new mentors through the FDL network join in to level up the project further. For example, an experienced industry engineer can perform code reviews, while an experienced scientist can review the experiments for the correctness and suggest other ways to amplify the value of the contributions. SpaceML provides access to experienced and often well-known professionals as volunteer mentors to the public who might not be accessible otherwise. To make efficient use of time for both the mentors and mentees, the mentees need to earn the time of mentors in the form of completed work, like through the early challenge or the weekly goals.

\textbf{Enhancing STEAM education}: {Distributed open-source research}, as spearheaded by SpaceML can work given (1) a specific focus for each contributor (2) a support structure of diverse experts and mentors (3) a plan detailing how all the individual contributions work together in the final deployment. Motivated by research with high public value that has a path towards deployment, high school and early graduate school students from a wide variety of socioeconomic backgrounds and geographies volunteered in the first iteration of SpaceML.

\section*{Impact of SpaceML}
In our pilot study, SpaceML contributors were high school students, with participants from Nigeria, Mexico, Korea, Germany, and the United States with no prior AI background and worked on tasks like self-supervised learning using SimCLR, multi-resolution image search, cost-efficient data labeling, balancing imbalanced data without labels, and generating synthetic patches for missing satellite imagery.

SpaceML has also boosted the monitoring and improvement of the AI model and web tool for the TRL-9 NASA CAMS project (The CAMS project, established in California in 2010, uses hundreds of low light CCTV cameras to capture the meteor activity in the night sky. After each night, the resident scientist performs the triangulation of tracklets captured by two or more cameras and computes the meteor's trajectory, orbit, and light curve. Each solution is then manually classified as a meteor or not a meteor like planes, birds, clouds etc. CAMS became a part of FDL in 2017 in order to automate the data processing pipeline and filtration of meteors.)

As part of FDL 2017, beyond the huge efficiency improvements from automation, an AI pipeline with human-level accuracy, and a new web tool \cite{2018pimo.conf...65D} that allowed global access to the previous night's activity, the FDL team’s contribution resulted in raising awareness and bringing in new citizen scientists who established stations in Brazil. The automation of the data processing allowed researchers to manage the camera stations better, which led to the {detection of the highest number of meteors in a single night (including 3003 Geminids \& 1154 sporadic meteors) in December 2017}. Subsequently, increased funding led to a four-fold global expansion of the camera network in the Southern Hemisphere.

As the CAMS project grew globally, requirements to add features to the application arose. During SpaceML 2020 a team of two students and industry mentors volunteered through open-source contributions to improve functionality. In 2020, the quick turnaround time of discovery from the CAMS web tool ultimately led to the an almost monthly reporting of a new and unusual meteor showers in the sky, including the {discovery of multiple meteor showers}, such as the gamma Piscis Austrinids and the sigma Phoenicids. One of those showers helped better define the orbit of parent comet Grigg-Mellish, which was observed poorly in 1907~\cite{JENNISKENS2020104979}.


\section*{Conclusion}
We present SpaceML, an open-source interdisciplinary science and AI research program. SpaceML releases challenges to the public inviting a diverse range of contributors. The released challenges are made accessible by packaging small projects within Google Colab (with free GPUs). Contributors who show promise are provided a personalized educational plan to build a foundation in the area of research and then receive relevant mentorship on the project from experts. Starting with freely available compute resources (eg Google Colab) contributors who show enough progress are provided cloud credits with high performance compute to allow for additional training and scaling. SpaceML has a focused aim of converting research into product reaching the highest NASA Technology Readiness Level (TRL 9). The opportunities afforded by SpaceML have brought in cohorts of citizen scientists and created opportunities for joint AI and science research. The first batch of SpaceML researchers were high school students who entered with no prior AI background,  pursued research in advanced topics like self-supervised learning on unlabelled satellite imagery, balancing unlabelled data, multi resolution search on satellite data, some eventually beat the state-of-the-art for certain techniques. The high school students also presented and/or published their work work in peer review conferences. Several of the high students who were participants in SpaceML have expressed interest in pursuing research in earth sciences, climate change and related STEAM fields as potential career options.


\bibliography{references}
\bibliographystyle{IEEEtran}

\medskip

\small

\end{document}